\documentclass[twoside]{article}
\usepackage[accepted]{aistats2022}
\usepackage{graphicx}
\usepackage{caption}
\usepackage{subcaption}
\usepackage[justification=centering]{caption}
\usepackage{courier}
\usepackage{blindtext}
\usepackage[table]{xcolor}
\usepackage[english]{babel}
\usepackage[utf8x]{inputenc}
\usepackage[skip=10pt]{caption} % example skip set to 2pt
\usepackage{ragged2e}

\usepackage{apacite}

\begin{document}



% \twocolumn[{
% \aistatstitle{Self-Supervised Visual Representation Learning Using Lightweight Architectures}

% \aistatsauthor{ Prathamesh Sonawane\textsuperscript{1}* \And Sparsh Drolia\textsuperscript{1}* \And Saqib Shamsi\textsuperscript{2} \And Bhargav Jain\textsuperscript{2} }
% \vspace{0.4cm}

% \centering
% \address{ 
% \textsuperscript{1}Pune Institute of Computer Technology, Maharashtra, India  \textsuperscript{2} Whirlpool Corporation\\
% pratt3000, sparshdrolia, shamsi.saqib)@gmail.com, bhargav\_jain@whirlpool.com
% }
% \vspace{0.6cm}
% }]%end twocolumn

\twocolumn[
\aistatstitle{Self-Supervised Visual Representation Learning Using Lightweight Architectures}

\aistatsauthor{ Prathamesh Sonawane\textsuperscript{1}* \And Sparsh Drolia\textsuperscript{1}* \And Saqib Shamsi\textsuperscript{2} \And Bhargav Jain\textsuperscript{2}}
\vspace{0.4cm}
\aistatsaddress{ 
\textsuperscript{1} Pune Institute of Computer Technology, Maharashtra, India  \textsuperscript{2} Whirlpool Corporation \\
(pratt3000, sparshdrolia, shamsi.saqib)@gmail.com, bhargav\_jain@whirlpool.com
}
]

\begin{abstract}
  In self-supervised learning, a model is trained to solve a pretext task, using a data set whose annotations are created by a machine. The objective is to transfer the trained weights to perform a downstream task in the target domain. We critically examine the most notable pretext tasks to extract features from image data and further go on to conduct experiments on resource constrained networks, which aid faster experimentation and deployment. We study the performance of various self-supervised techniques keeping all other parameters uniform. We study the patterns that emerge by varying model type, size and amount of pre-training done for the backbone as well as establish a standard to compare against for future research. We also conduct comprehensive studies to understand the quality of representations learned by different architectures.
\end{abstract}

\section{INTRODUCTION}
Self-supervised learning is a class of methods that leverage \textit{pretext} tasks to create labels automatically from the data. These labels are used to learn representations. The representations learned from the pretext task are then used to train the same model on some task of interest called the \textit{downstream} task. Self-supervised learning is prominently used in various tasks including NLP(Natural Language Processing), computer vision and reinforcement learning to learn spatial and temporal features depending on the task, data set and model architecture. 

Although we have observed significant progress in Self-Supervised learning research, it still remains to be outperform the traditional way of training models which is by using human annotated data sets. Although in recent research Self-Supervised learning techniques have come close to performing on par with Supervised training, they appear somewhat unapproachable due to the amount of compute as well as time demanded for experimentation and research. Despite its said drawback, it does provide a way of eliminating label acquisition cost and reducing time required to create a data set. Moreover the networks trained aren't hinged on to a specific task like classification or object detection, the models weights remain universal for a variety of downstream tasks.

It is to these benefits that we are observing an increased amount of research in similar fields like transfer learning, semi-supervised learning, weakly supervised learning and unsupervised learning. In this paper we focus on self-supervised learning. We study the most notable pretext tasks for visual representation learning on lightweight architectures in a resource constrained environment.

Precisely, we aim to answer the rather intriguing question, \textit{"How well do techniques like Rotnet \cite{Rotnet}, BYOL \cite{BYOL}, SimCLR \cite{SimCLR}, etc., which are traditionally trained on computationally heavier architectures like AlexNet\cite{alexnet}, ResNet \cite{resnet} learn features when they are trained on lightweight architectures like an EfficientNet-lite0 \cite{effnetlite}?"} We conduct a comparative analysis across various techniques with a controlled set of hyper parameters, and see how they perform in a low resource setting. We hope to establish a standard against which future research can be conducted and evaluated prematurely. Our aim is to establish a standard for comparison before dedicating huge chunks of time and capital in order to perform  pretext tasks on bigger architectures and consequently bigger data sets. Additionally in our study we also found that lightweight architectures could achieve comparable accuracies with significantly less carbon footprints(Tab. \ref{tab:top-validation-acc}).

To measure the computation complexity we have not only used the number of Floating Point Operations (FLOPs) performed by each model, as it is an indirect metric, but also have measured amount of training time, which is a more direct measure \cite{FLOPSvsTIME}. You can find all of the code for our experiments here.[Once the the paper gets accepted we will release the link to our code base here]

\section{RELATED WORK}
With the success of deep neural networks, a lot of tasks can be solved very well by collecting a labelled data set and using supervised learning. In order to perform well, the models usually need a large corpus of labeled examples. However, getting labels for data turns out to be expensive and scaling it up is a major challenge. With the vast amount of unlabeled data being generated in the form of text, images, videos,  everyday, the goal of self-supervised learning is to get supervision from the data itself to learn useful representations instead of using explicit labels. Once a good representation is learned on a \textit{pretext} task, the model can be fine tuned on a \textit{downstream} task with comparatively fewer data than what would have been required were the model trained from scratch on the downstream task.

The general nature of self-supervised learning allows it to be used across different modalities including images, text, video, audio and even in robotics. The use of context to predict words \cite{word2vec,glove,ulmfit,bert} is a popular technique in the domain of natural language processing. In the similar vein, the temporal information in video can also be leveraged to learn representations \cite{shuffle_and_learn,arrow_of_time}. One could also use different modalities in videos for the same \cite{cross_and_learn,multimodal}.

There has been a lot of work over the years in the field of visual representation learning. Doersch et al. \cite{patches_ssl} proposed a patch based method to predict the placement of patches relative to each other within an image. They motivated a line of patch based methods such as the "jigsaw puzzle" based task \cite{jigsaw}, where they used nine patches from the full image instead of just two patches to make the pretext task more challenging. There have been more works following the two \cite{jigsaw++,learning_to_count}.

In contrast to patch based methods, there have been other methods that used the image information to create a classification pretext task. Notable examples of these include RotNet \cite{Rotnet}, where the authors rotate the images in multiples of 90 degrees and the model performs a 4-way classification predicting the angle the image was rotated by. Another class of methods have focused on generative modeling to learn representations from images. Researchers have relied on tasks like predicting a subset of channels of the image from another subset of the same image such as in a colorization pretext task \cite{colorful_colorization} and its improvement Split-Brain Auto Encoders \cite{SplitBrainAEZhang2017}. Pathak et al. used image in painting as a pretext task for self-supervised learning \cite{inpainting}.

Recently there has been a rise in the use of contrastive methods which have performed extremely well in this domain. Contrastive methods aim to learn a model which transforms an input into an embedding vector such that examples from the same class have similar embeddings relative to embeddings of samples from different classes. SimCLR \cite{SimCLR} is a framework which learns representations by maximizing the agreement between different augmented views of the same input in a latent space. Barlow Twins \cite{barlow_twins} method learns to make the cross-correlation matrix between the two distorted versions of the same image close to the identity matrix. MOCO \cite{moco} and MOCO-v2 \cite{mocov2} rely on a framework of representation learning from images as a dynamic dictionary look-up.  BYOL \cite{BYOL} aims to learn a representation using two neural networks, which are referred to as online and target networks, without the use of negative samples. The two networks have the same architecture, with the target network having polyak averaged weights.

There has been a lot of focus and effort from the research community on coming up with new pretext tasks. We aim to take a complementary approach in the self-supervised research landscape by evaluating various pretext tasks on low resource architectures like MobileNetv2 \cite{mobilenetv2}, ShuffleNetv2 \cite{shufflenetv2}, SqueezeNet \cite{squeezenet} and EfficientNetLite0 \cite{effnetlite} and compare and contrast the performance on a computationally expensive architecture like a ResNet \cite{resnet}. Our work is similar to the study by Kolesnikov et al. \cite{revisiting_ssl} where they investigate how architectural choices affect the performance of various pretext tasks for visual representation learning. However, while they use different variants of Residual Networks \cite{resnet}, which are computationally expensive architectures, the aim of our work is to examine how self-supervised visual representation learning fares on computationally lighter architectures. 

\begin{table*}[ht]
\centering
\caption{Max validation accuracy achieved for downstream classification task on different pretext tasks and backbone architectures. C.T. refers to contrastive techniques. The column mentions if the respective technique is contrastive or not. CO$_2$ Gen. is CO$_2$ generated, as in the carbon emitted in kg CO$_2$ equivalent for training on average for the respective architecture.}
\label{tab:top-validation-acc}
\begin{tabular}{cccccccc}
\cline{1-6} \cline{8-8}
\multicolumn{1}{|c|}{\cellcolor[HTML]{9B9B9B}\textbf{Val. Acc.}} & \multicolumn{1}{c|}{\cellcolor[HTML]{C0C0C0}\textbf{Efficientnetlite0}} & \multicolumn{1}{c|}{\cellcolor[HTML]{C0C0C0}\textbf{Mobilenetv2}} & \multicolumn{1}{c|}{\cellcolor[HTML]{C0C0C0}\textbf{ResNet-18}} & \multicolumn{1}{c|}{\cellcolor[HTML]{C0C0C0}\textbf{Shufflenetv2}} & \multicolumn{1}{c|}{\cellcolor[HTML]{C0C0C0}\textbf{Squeezenet}} & \multicolumn{1}{c|}{.} & \multicolumn{1}{c|}{\cellcolor[HTML]{C0C0C0}\textbf{C.T.}} \\ \cline{1-6} \cline{8-8} 
\multicolumn{1}{|c|}{\cellcolor[HTML]{C0C0C0}\textbf{Rotnet}} & \multicolumn{1}{c|}{\cellcolor[HTML]{FFFFFF}\textbf{0.28}} & \multicolumn{1}{c|}{0.31} & \multicolumn{1}{c|}{0.33} & \multicolumn{1}{c|}{0.37} & \multicolumn{1}{c|}{0.30} & \multicolumn{1}{c|}{} & \multicolumn{1}{c|}{No} \\ \cline{1-6} \cline{8-8} 
\multicolumn{1}{|c|}{\cellcolor[HTML]{C0C0C0}\textbf{MOCOv2}} & \multicolumn{1}{c|}{0.47} & \multicolumn{1}{c|}{0.51} & \multicolumn{1}{c|}{\cellcolor[HTML]{FFFFFF}\textbf{0.78}} & \multicolumn{1}{c|}{\cellcolor[HTML]{FFFFFF}0.65} & \multicolumn{1}{c|}{\cellcolor[HTML]{FFFFFF}0.10} & \multicolumn{1}{c|}{} & \multicolumn{1}{c|}{Yes} \\ \cline{1-6} \cline{8-8} 
\multicolumn{1}{|c|}{\cellcolor[HTML]{C0C0C0}\textbf{Split Brain}} & \multicolumn{1}{c|}{0.58} & \multicolumn{1}{c|}{0.57} & \multicolumn{1}{c|}{0.59} & \multicolumn{1}{c|}{0.53} & \multicolumn{1}{c|}{0.45} & \multicolumn{1}{c|}{} & \multicolumn{1}{c|}{No} \\ \cline{1-6} \cline{8-8} 
\multicolumn{1}{|c|}{\cellcolor[HTML]{C0C0C0}\textbf{SimCLR}} & \multicolumn{1}{c|}{0.69} & \multicolumn{1}{c|}{0.65} & \multicolumn{1}{c|}{0.70} & \multicolumn{1}{c|}{0.59} & \multicolumn{1}{c|}{0.60} & \multicolumn{1}{c|}{} & \multicolumn{1}{c|}{Yes} \\ \cline{1-6} \cline{8-8} 
\multicolumn{1}{|c|}{\cellcolor[HTML]{C0C0C0}\textbf{BYOL}} & \multicolumn{1}{c|}{0.51} & \multicolumn{1}{c|}{0.71} & \multicolumn{1}{c|}{\cellcolor[HTML]{FFFFFF}\textbf{0.78}} & \multicolumn{1}{c|}{0.71} & \multicolumn{1}{c|}{0.68} & \multicolumn{1}{c|}{} & \multicolumn{1}{c|}{Yes} \\ \cline{1-6} \cline{8-8} 
. &  &  &  &  &  &  &  \\ \cline{1-6}
\multicolumn{1}{|c|}{\cellcolor[HTML]{C0C0C0}\textbf{Avg. FLOPs}} & \multicolumn{1}{c|}{3.90E+08} & \multicolumn{1}{c|}{3.27E+08} & \multicolumn{1}{c|}{1.80E+09} & \multicolumn{1}{c|}{4.01E+07} & \multicolumn{1}{c|}{3.20E+08} &  &  \\ \cline{1-6}
\multicolumn{1}{|c|}{\cellcolor[HTML]{C0C0C0}\textbf{CO2 Gen.}} & \multicolumn{1}{c|}{1.9} & \multicolumn{1}{c|}{1.8} & \multicolumn{1}{c|}{3.22} & \multicolumn{1}{c|}{1.84} & \multicolumn{1}{c|}{1.99} &  &  \\ \cline{1-6}
\end{tabular}
\end{table*}

\section{EXPERIMENTAL SETUP}
\label{section:experimental_setup}
\subsection{Dataset}We have used STL-10 \cite{STL10} for all of our experiments. STL-10 is made up of a subset of images from the ImageNet \cite{imagenet} . It is composed of labeled and unlabeled subsets. The first set consists of $10$ classes with $500$ training and $800$ validation RGB images per class. The images are of size $96 \times 96$. These are used for downstream training and evaluation. The second set consists of $100,000$ unlabelled RGB images of size $96 \times 96$, which are used for pretext task training.

% graphs 1
\begin{figure*}[ht]

\centering
\begin{subfigure}{.32\textwidth}
  \centering
  % include first image
  \includegraphics[width=\linewidth]{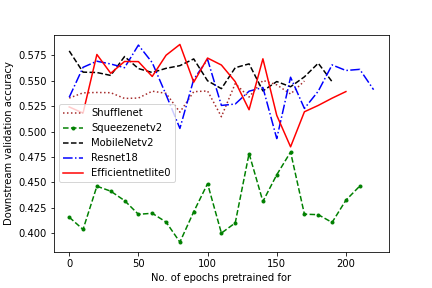}  
  \caption{Split-Brain Auto Encoders}
  \label{fig:val_acc_graphs1}
\end{subfigure}
\begin{subfigure}{.32\textwidth}
  \centering
  % include second image
  \includegraphics[width=\linewidth]{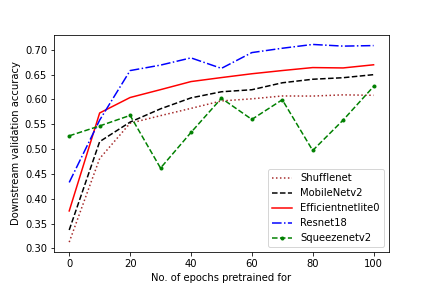}  
  \caption{SimCLR}
  \label{fig:val_acc_graphs2}
\end{subfigure}
\begin{subfigure}{.32\textwidth}
  \centering
  % include second image
  \includegraphics[width=\linewidth]{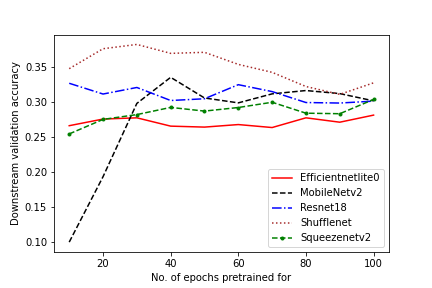}  
  \caption{Rotnet}
  \label{fig:val_acc_graphs3}
\end{subfigure}

\begin{subfigure}{.32\textwidth}
  \centering
  % include third image
  \includegraphics[width=\linewidth]{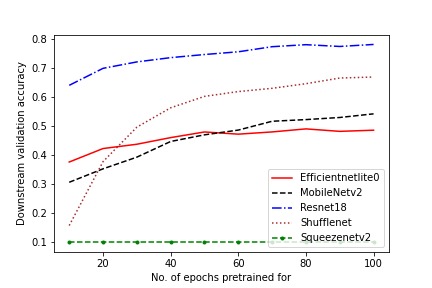}  
  \caption{MOCOv2}
  \label{fig:val_acc_graphs4}
\end{subfigure}
\begin{subfigure}{.32\textwidth}
  \centering
  % include fourth image
  \includegraphics[width=\linewidth]{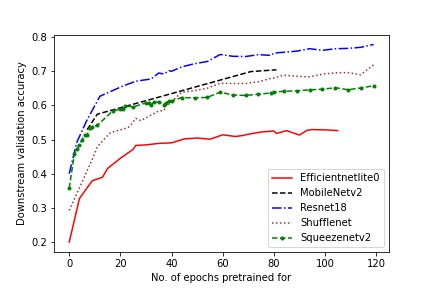} 
  \caption{BYOL}
  \label{fig:val_acc_graphs5}
\end{subfigure}
\caption{The graphs represent max validation accuracies achieved on downstream classification task with increased amount of pretraining done(in intervals of 10 epochs) on different architectures for every technique. Each graph represents one among 5 techniques evaluated. (x-axis : number of epochs for which the model was pre-trained. y-axis : Max validation accuracy achieved on downstream classification task.)
\label{fig:val_acc_graphs}}
\end{figure*}

\subsection{Architectures}
\label{subsection:architectures}
\subsubsection{ResNet-18}
To compare the results of the different techniques against a computationally expensive architecture we train a ResNet-18 \cite{resnet} on all of the techniques separately. The network consists of an initial $7 \times 7$ convolutional layer followed by a max pooling operation. The network then consists of $16$ $3 \times 3$ convolutions and ReLU non-linearity. Down sampling is performed by a strided convolutions instead of pooling layers. The network has $1.8$ GFLOPs (multiply-adds).
% Note: compare the percentage of FLOPs in the other architectures to this as the resnet 
% authors do it in the baseline section.

\subsubsection{Mobilenet-v2} 
We chose Mobilenet-v2 \cite{mobilenetv2} as one of the lightweight architectures for this study. Mobilenet-v2 is built using inverted residual blocks with shortcut connections between thin bottleneck layers. Its design includes an initial fully convolutional layer with 32 filters, followed by 19 residual bottleneck layers. The size of all the kernels in all of the convolution layers is $3 \times 3$

\subsubsection{Efficientnet-lite0}
We compared the performance of Efficientnet-lite0 \cite{effnetlite} as it is one of the modern state-of-the-art architectures. It is a smaller version of EfficientNet \cite{effnetlite} and uses compound scaling to uniformly scale depth $\alpha$,  width $\beta$, image size $\gamma$. Neural architecture search is used to get the baseline model and bigger architectures are obtained by scaling this up. It consist of convolutions followed by inverted residual blocks found in MobileNet-v2. 

\subsubsection{Squeezenet} 
We also trained Squeezenet \cite{squeezenet} in addition to the architectures mentioned above on all the techniques separately. It is composed of fire modules that further consists of a squeezed convolution layer (has only $1 \times 1$ filters), that feeds into an expand layer (that has a mix of $1 \times 1$ and 3×3 convolution filters). The architectures is constructed using a convolution layer, followed by $9$ fire modules, followed by a convolution and a Softmax unit at the end. The down sampling in SqueezeNet is performed by pooling layers.

\subsubsection{Shufflenet-v2} 
Shufflenet-v2 \cite{shufflenetv2} was our final lightweight architecture for this study. Shufflenet-v2 unit is a residual block. In its residual branch, for the $3 \times $3 layer, a computational economical $3 \times $3 depth wise convolution on the bottleneck feature map is applied. Then the first $1 \times $1 layer is replaced with point wise group convolution followed by a channel shuffle operation, to form a Shufflenet unit. For example, given the input size c x h x w and the bottleneck channels m, Shufflenet unit requires only h*w(2c*m/g + 9m) FLOPs, where g means the number of groups for convolutions.

\subsection{Techniques}
\subsubsection{Rotnet} 
The authors proposed rotating an image pseudo-randomly by a discrete set of angles \cite{Rotnet}. The angle is treated as the label and the rotated image as the input. In the paper they tested rotating images in 45, 90 and 180 degree intervals separately. The best results were observed for 90 degree interval set and the worst results were for 180 degree interval set. We used 90 degree intervals as our labels for training since they achieved the highest accuracy in the paper.

\subsubsection{MOCOv2} 
The authors introduced a contrastive learning approach – momentum contrast (MOCO) \cite{mocov2}. The key ideas of this method are: 1) implementation of a queue as the dictionary which stores plethora of keys; 2) updating the key encoder using the momentum update from the query encoders(without the need of passing the batch through key encoder). Compared with previous contrastive learning methods based on memory bank and end-to-end learning, MOCOv2 not only supports a large negative sample size but also maintains a consistent key encoding.

\subsubsection{SimCLR}
The authors introduced a framework \cite{SimCLR}  for contrastive learning of visual representations. SimCLR learns representation by maximising similarity of two differently augmented data set of the same image. It uses a stochastic data augmentation module that randomly modifies a given data instance, producing two correlated images of the same example, which is regarded a positive pair. SimCLR uses three basic augmentations in sequence: random cropping followed by resizing to the original size, random colour distortions, and random Gaussian blur. The authors believe that random cropping and colour distortion are critical to achieving high performance.

% graphs 2
\begin{figure*}[ht]
\centering
\begin{subfigure}{.32\textwidth}
  \centering
  % include first image
  \includegraphics[width=\linewidth]{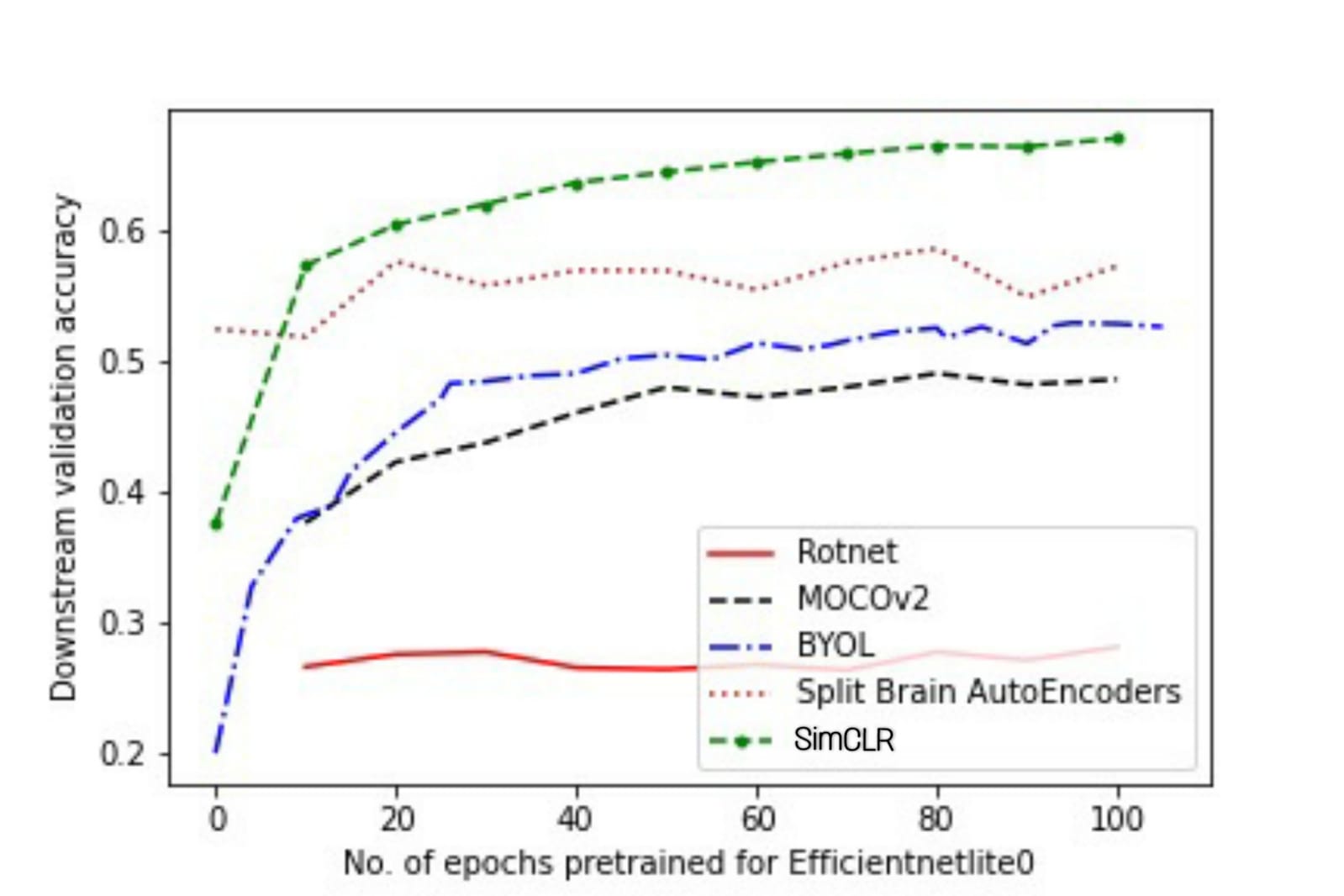}  
  \caption{Efficientlite0}
  \label{fig:arch_wise_graphs1}
\end{subfigure}
\begin{subfigure}{.32\textwidth}
  \centering
  % include second image
  \includegraphics[width=\linewidth]{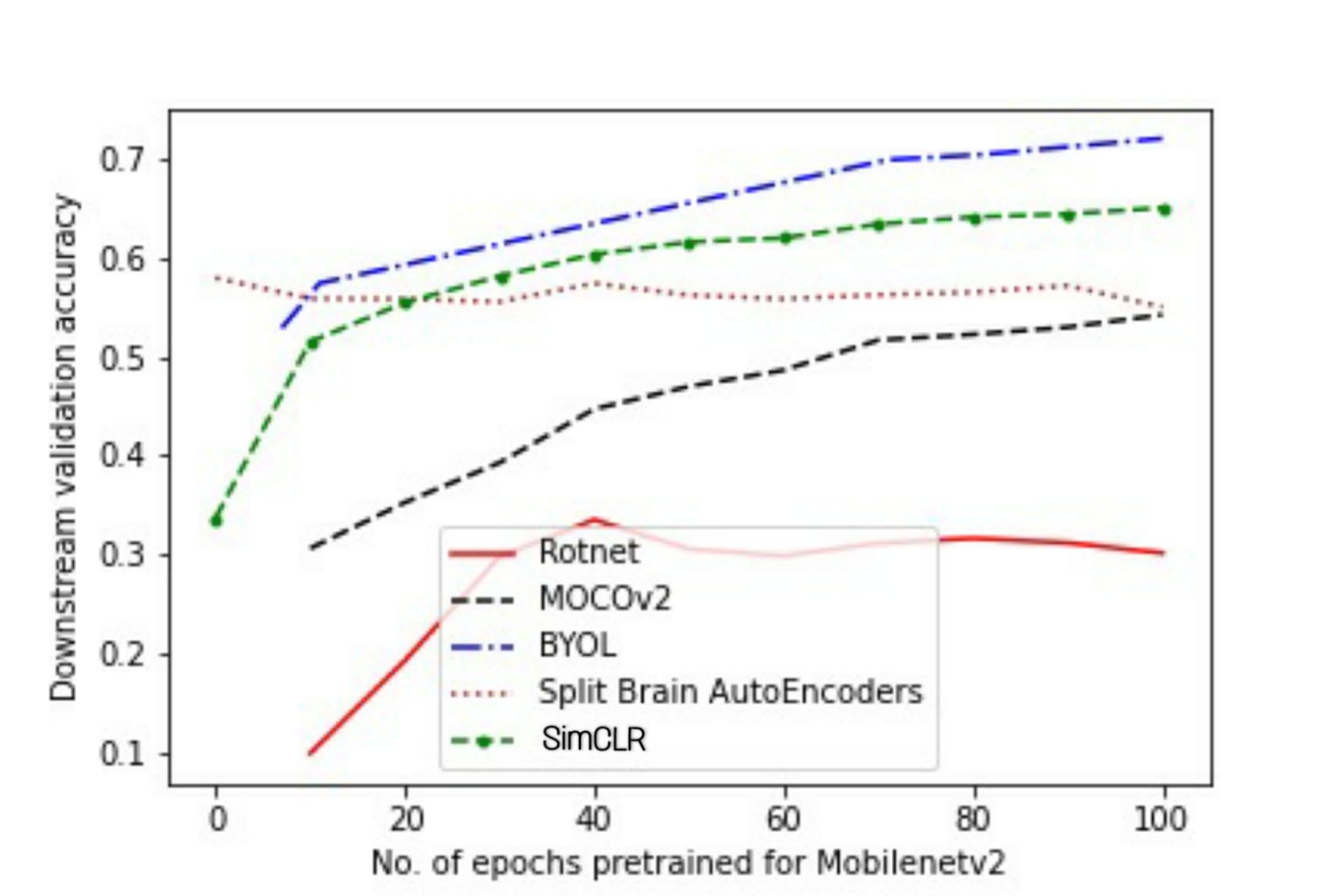}  
  \caption{Mobilenetv2}
  \label{fig:arch_wise_graphs2}
\end{subfigure}
\begin{subfigure}{.32\textwidth}
  \centering
  % include second image
  \includegraphics[width=\linewidth]{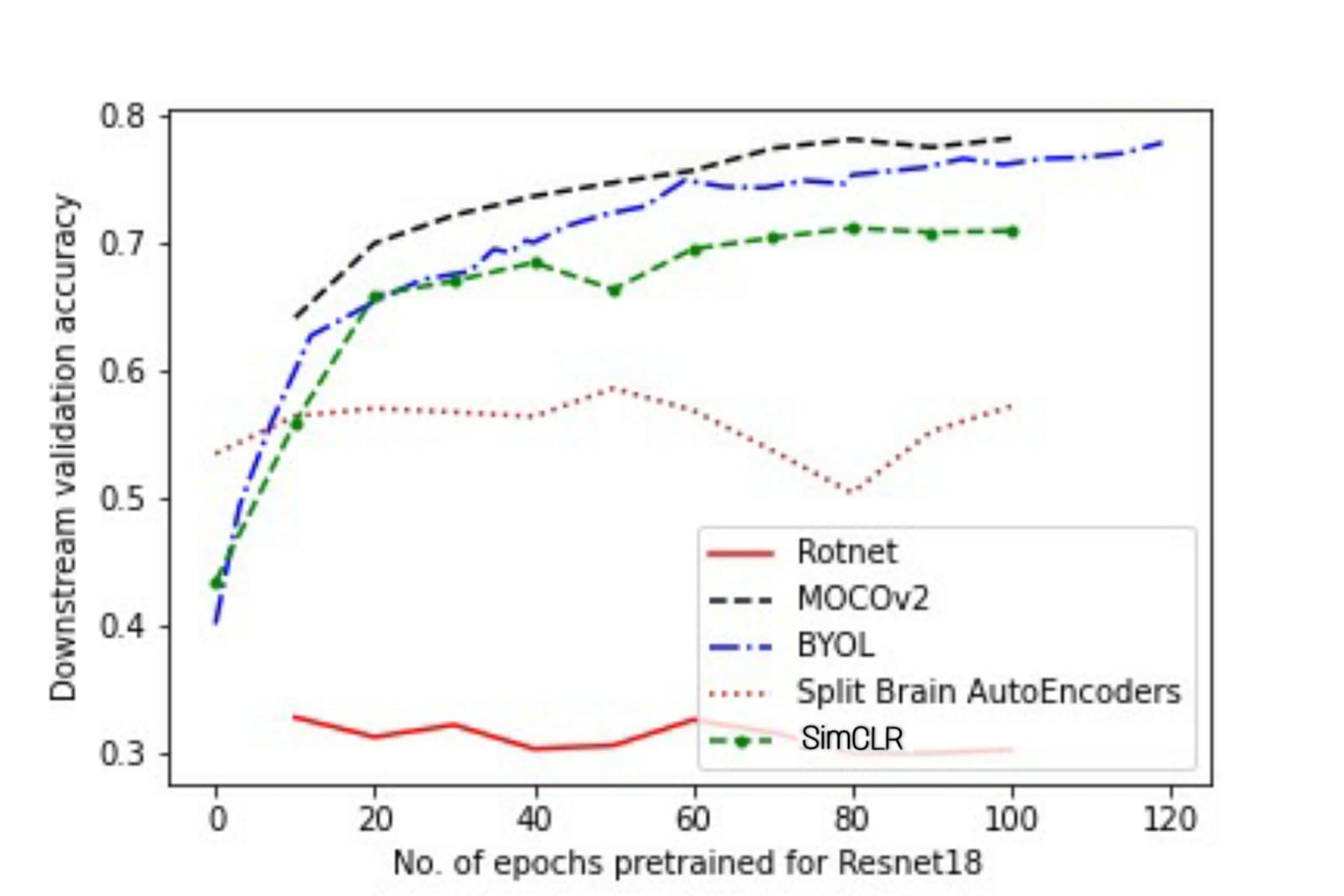}  
  \caption{ResNet-18}
  \label{fig:arch_wise_graphs3}
\end{subfigure}

\begin{subfigure}{.32\textwidth}
  \centering
  % include third image
  \includegraphics[width=\linewidth]{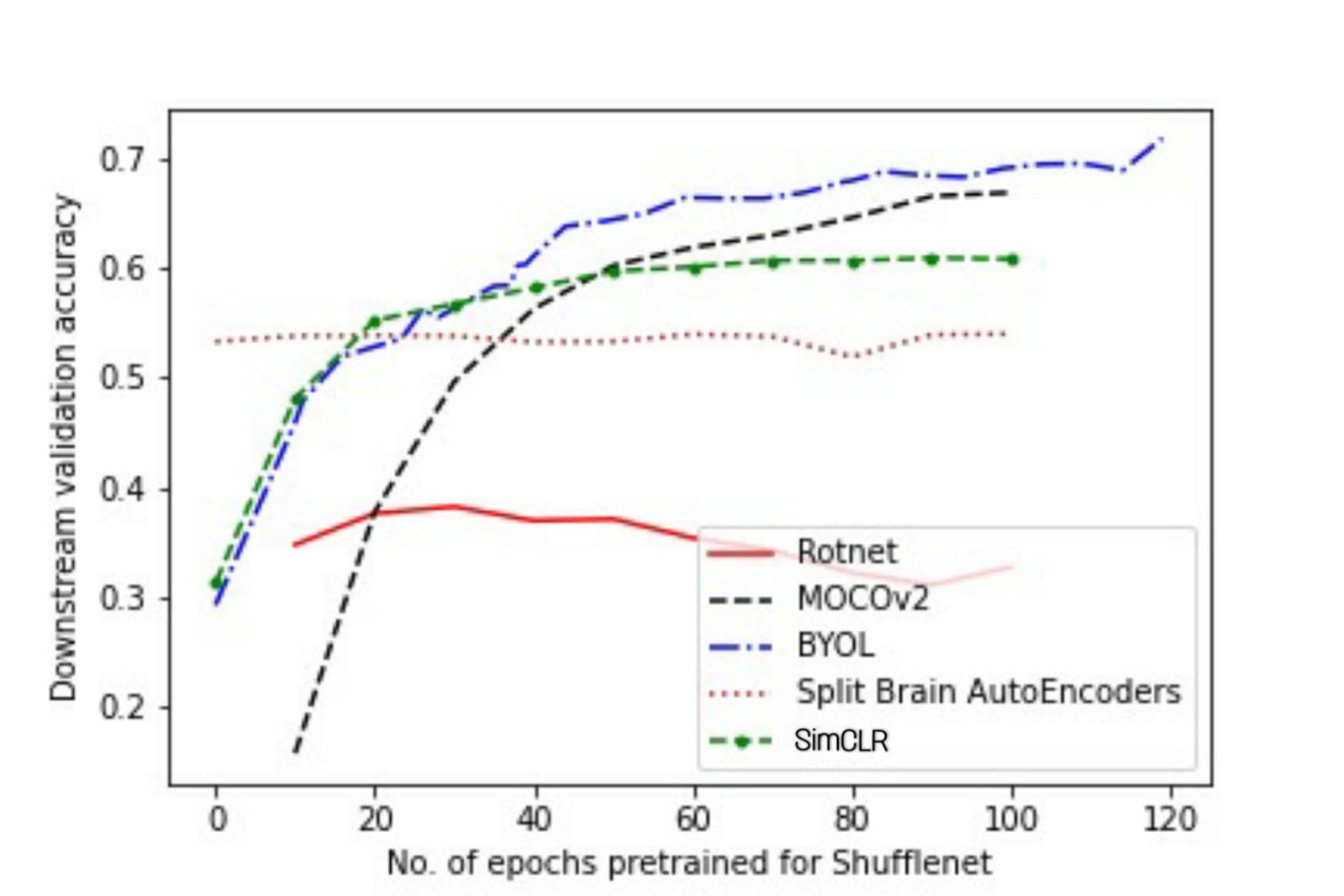}  
  \caption{Shufflenet}
  \label{fig:arch_wise_graphs4}
\end{subfigure}
\begin{subfigure}{.32\textwidth}
  \centering
  % include fourth image
  \includegraphics[width=\linewidth]{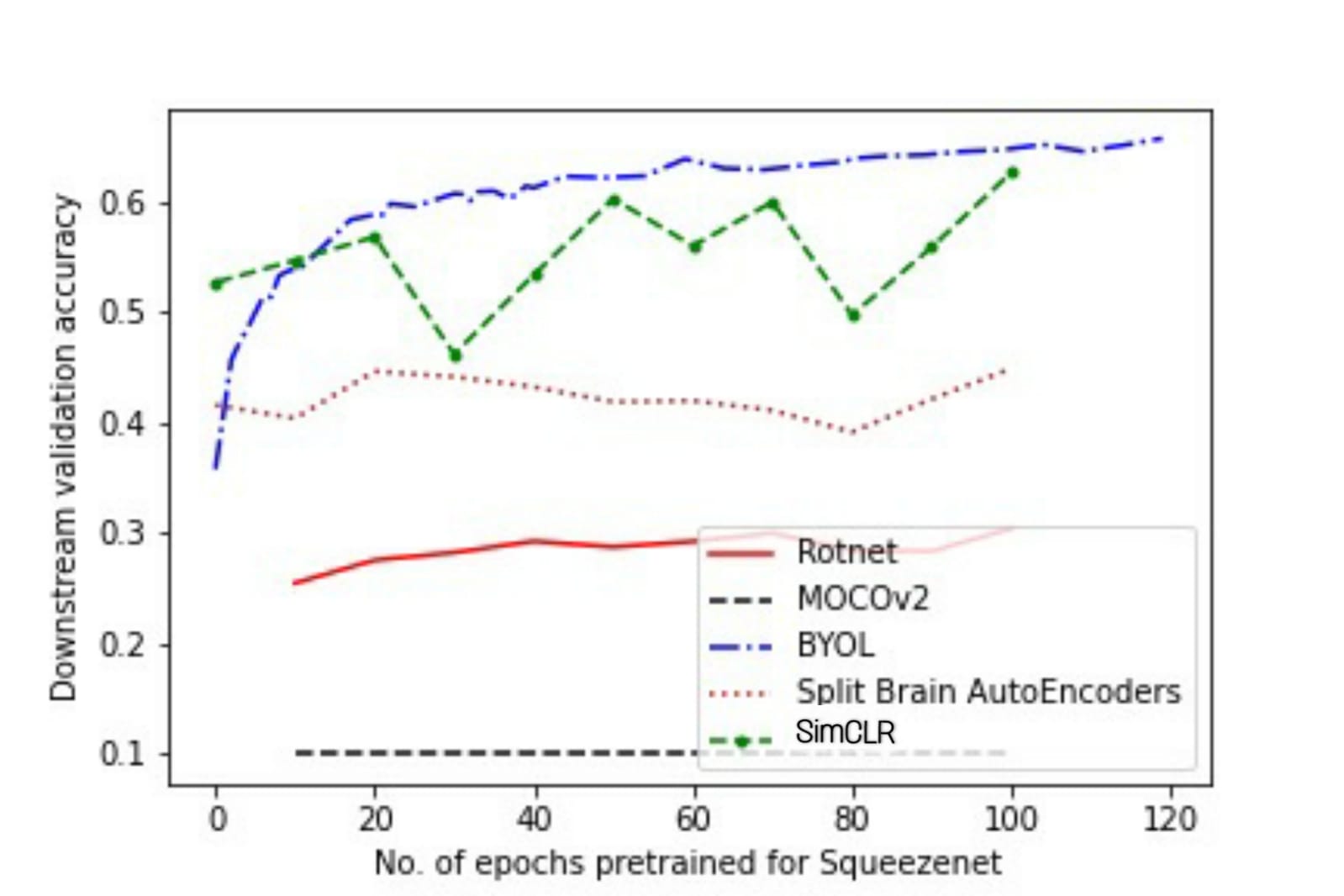}  
  \caption{Squeezenet}
  \label{fig:arch_wise_graphs5}
\end{subfigure}

\caption{The graph represents max validation accuracies achieved on downstream classification task with the increase in pretraining done on different techniques for each architecture. Each graph represents one among 5 architectures evaluated.(x-axis : number of epochs for which the model was pre-trained. y-axis : Max validation accuracy achieved on downstream classification task.)}
\label{fig:arch_wise_graphs}
\end{figure*}

\subsubsection{BYOL}
BYOL \cite{BYOL} is a contrastive learning approach which uses two encoder networks with the same architecture referred to as online and target networks to learn representations and reduces the contrastive loss between the representations learned by the two networks. Unlike other contrastive learning BYOL does not use any negative samples. The output of the network is iterated so as to serve as a target for enhanced representation. BYOL trains its online network using an augmented view of an image and predicts the target network’s representation of another augmented view of the same image. They use ResNet-50 as an encoder network. For the projection MLP, a 2048 dimensional feature vector is projected onto 4096-dimensional vector space first with a sub-network composed of batch norm followed by ReLU non-linear activation. The resulting vector is then reduced to a 256-dimensional feature vector. The same architecture is used for the predictor network. 

\subsubsection{Split-Brain Auto encoders}
Split-Brain Auto encoders are a spin on the traditional auto encoder architecture \cite{SplitBrainAEZhang2017}. The method adds a split to the network thereby creating two sub-networks. The image is divided into two subsets of channels. Each sub-network predicts one subset of channels from the other subset. For this study, we use the \textit{Lab} space and divide the image into perceptual lightness \textit{L} and color \textit{ab} channels. Both sub-networks are trained for classification using a cross-entropy objective. When predicting \textit{L} from \textit{ab}, the output space is quantized into 50 bins of size 2. When predicting \textit{ab} from \textit{L}, the quantized output space is binned into 313 bins of size 100.

\section{EXPERIMENTS}
We evaluated the performance of five of the most cited techniques: Rotnet, MOCOv2, Split-Brain Auto encoders, SimCLR and BYOL across five different architectures: Efficientnetlite0, Mobilenetv2, ResNet-18, Shufflenetv2 and Squeezenet. All of them are relatively smaller architectures, except for ResNet-18 which acts as a point of comparison of the performance of the four lightweight architectures. All of the techniques and architectures have been described in Section \ref{section:experimental_setup} above.

Every architecture was trained on the pretext task, using the unlabelled subset of images of STL-10 until convergence. We saved the weights of the network every 5 epochs during pretraining. These checkpoints were later used to evaluate performance on the downstream classification task to evaluate the performance as a function of the amount of pre-training the networks were exposed to for the pretext task.

We used categorical crossentropy loss for the downstream training and Adam \cite{adam} as the optimizer. For pretext tasks all of the configuration was kept as stated in their respective papers, only the model architectures were replaced.

To inspect the quality of the learned representations and understand what part of the images most affect the outcome of predictions made by the models, we used Saliency maps \cite{Saliency}. Saliency maps show us the degree of importance of each pixel an image, in the visual field. Our approach to this was to find the best true positives and false negatives for each experiment and then examining those for any observable trends.

Another technique we used to understand the models' performance is K-Nearest Neighbours search (kNN) \cite{KNN}. We use it to find the closest image representations in space of representations learned by the models. We use Euclidean distance as our distance measure.

% Saliency 
\begin{figure*}[ht]
\centering
\begin{subfigure}{.28\textwidth}
  \centering
  % include first image
  \includegraphics[width=\linewidth]{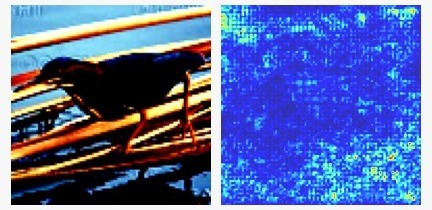}  
  \caption{Bird}
  \label{fig:Sal_Bird}
\end{subfigure}
\hspace{1em}
\begin{subfigure}{.28\textwidth}
  \centering
  % include second image
  \includegraphics[width=\linewidth]{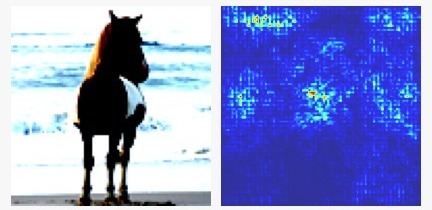}  
  \caption{Horse}
  \label{fig:Sal_Horse}
\end{subfigure}
\hspace{1em}
\begin{subfigure}{.28\textwidth}
  \centering
  % include second image
  \includegraphics[width=\linewidth]{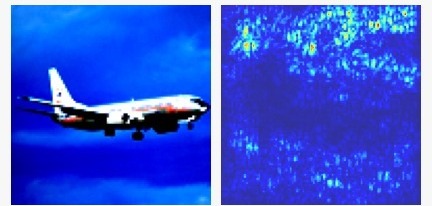}  
  \caption{Aeroplane}
  \label{fig:Sal_Plane}
\end{subfigure}
\hspace{1em}
\begin{subfigure}{.28\textwidth}
  \centering
  % include third image
  \includegraphics[width=\linewidth]{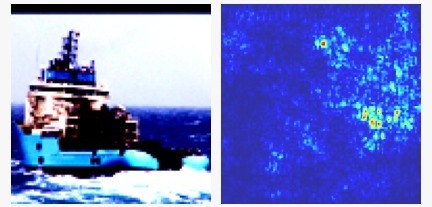}  
  \caption{Ship}
  \label{fig:Sal_ship}
\end{subfigure}
\hspace{1em}
\begin{subfigure}{.28\textwidth}
  \centering
  % include fourth image
  \includegraphics[width=\linewidth]{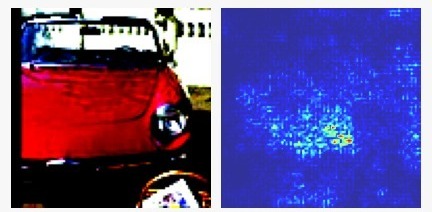}  
  \caption{Car}
  \label{fig:Sal_truck}
\end{subfigure}
% #scale
\\
\hspace{1em}
\begin{subfigure}{.28\textwidth}
  \centering
  % include fourth image
  \includegraphics[width=\linewidth]{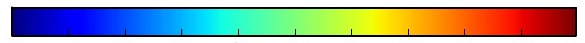}  
  \caption{Saliency heat map scale. Blue showing the least focused regions and red showing the most focused regions}
  \label{fig:sal_scale}
\end{subfigure}

\caption{Saliency maps for BYOL using Mobilenetv2's backbone. Although only a few examples have been shown, These Saliency maps represent a trend which we observed in most of our experiments. In Fig \ref{fig:Sal_Bird} we can observe that the model focuses on the surrounding rather than the bird and hence throughout our experiments bird images were the most misclassified. In Fig \ref{fig:Sal_Horse} we can observe that the model focuses on the horse with a decent regard for its surrounding as well, this must have been the reason that horse images were classified correctly in most instances. In Fig \ref{fig:Sal_Plane} and Fig \ref{fig:Sal_ship} we can observe that the model focuses primarily on the blue pixels, and thus ends up confusing between planes and ships in practice. In Fig \ref{fig:Sal_truck} we can see that the model is focusing on the red pixels and hence we get an explanation for why red coloured cars and trucks are misclassified interchangeably by models   }
\label{fig:saliencymaps}
\end{figure*}

% \vspace{-9mm}
\section{Results}
\subsection{General}
As is evident from figures \ref{fig:val_acc_graphs} Resnet-18 is the best performer in Fig \ref{fig:val_acc_graphs2}, Fig \ref{fig:val_acc_graphs4} and Fig \ref{fig:val_acc_graphs5}, with the different in performance increasing with the amount of pretraining. This was expected as it is significantly heavier than all architectures, . The highest validation accuracy in all our experiments was achieved using ResNet-18, twice (Table \ref{tab:top-validation-acc}). For Split-Brain Auto encoders and Rotnet, ResNet-18 has the second best performance. Split-Brain appears to be unstable when trained on ResNet-18 (as we can see from Fig \ref{fig:val_acc_graphs1}). For Rotnet, after 100 epochs of pretraining all architectures have about the same validation accuracy (Fig \ref{fig:val_acc_graphs3}), so it is unclear as to which is the best performing architecture.  Despite all this it is safe to say that ResNet-18 outperforms lighter models overall. On the other hand we can also observe in Fig \ref{fig:val_acc_graphs} that Squeezenet appears consistently in the worst 2 performing models. 

Another key take away from Fig \ref{fig:val_acc_graphs4} and Fig \ref{fig:val_acc_graphs5} is that while Shufflenet has trouble learning with less pretraining, it eventually overtakes the other architectures as the amount of pretraining increases. Hence, while Shufflenetv2 has a slower learning pace, with respect to the pretraining done, it learns better feature representations over time, unlike the other 4 architectures. 

We can observe from Fig \ref{fig:arch_wise_graphs} that BYOL is the best performing technique for 4 out 5 model architectures. From fig \ref{fig:val_acc_graphs} we can observe that it also is the most stable among other techniques. 

Split-Brain Auto Encoders and Rotnet require relatively less pretraining to reach their max validation accuracies, regardless of the architecture ( Fig \ref{fig:val_acc_graphs1} and Fig \ref{fig:val_acc_graphs3}). Rotnet is stable while doing so but the same cant be said about Split-Brain Autoencoders.

Another observation is that excessive pre-training didn't have any detrimental or supplemental effects on performance during the downstream task. 
% Initially we experimented with pre-training for upto 500 epochs during experiments but since no notable increments were observed after about 100 epochs we decreased the maximum pretraining epoch limit.

\subsection{Saliency Maps and kNN}
\begin{figure*}[ht]
\centering
% #1
\begin{subfigure}{.08\textwidth}
  \centering
  % include first image
  \includegraphics[width=\linewidth]{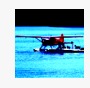}
    \caption{Plane}
  \label{fig:KNN_Plane_main}
\end{subfigure}
\hspace{3em}
\begin{subfigure}{.62\textwidth}
  \centering
  % include second image
  \includegraphics[width=\linewidth]{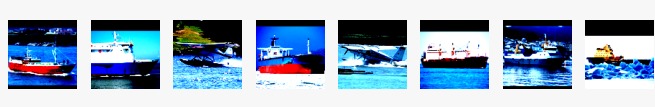} 
  \caption{nearest neighbour results for Shufflenetv2 trained with BYOL as the pretext task.}
  \label{fig:KNN_Plane_rest}
\end{subfigure}

% #2
\begin{subfigure}{.08\textwidth}
  \centering
  % include first image
  \includegraphics[width=\linewidth]{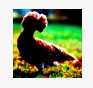} 
  \caption{Bird}
  \label{fig:KNN_Bird_main}
\end{subfigure}
\hspace{3em}
\begin{subfigure}{.62\textwidth}
  \centering
  % include second image
  \includegraphics[width=\linewidth]{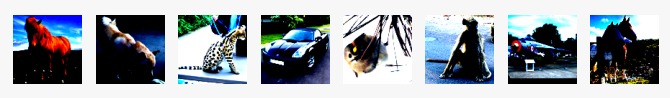} 
  \caption{Nearest neighbour results SqueezeNet trained with BYOL as the pretext task.}
%   \caption{Plane}
  \label{fig:KNN_Bird_rest}
\end{subfigure}

% #3
\begin{subfigure}{.08\textwidth}
  \centering
  % include first image
  \includegraphics[width=\linewidth]{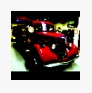} 
  \caption{Truck}
  \label{fig:KNN_Car_main}
\end{subfigure}
\hspace{3em}
\begin{subfigure}{.62\textwidth}
  \centering
  % include second image
  \includegraphics[width=\linewidth]{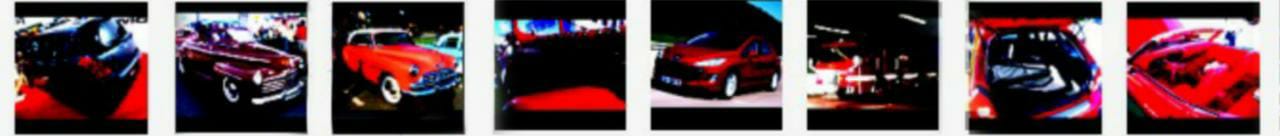} 
%   \caption{Plane}
\caption{Nearest neighbour results for Mobilenetv2 trained with BYOL as the pretext task.}
  \label{fig:KNN_Car_rest}
\end{subfigure}

% #4
\begin{subfigure}{.08\textwidth}
  \centering
  % include first image
  \includegraphics[width=\linewidth]{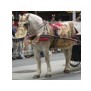}
  \caption{Horse}
  \label{fig:KNN_Horse_main}
\end{subfigure}
\hspace{3em}
\begin{subfigure}{.62\textwidth}
  \centering
  % include second image
  \includegraphics[width=\linewidth]{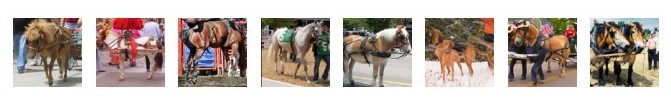} 
%   \caption{Plane}
\caption{Nearest neighbour results for Efficientnetlite0 trained with BYOL as the pretext task.}
  \label{fig:KNN_Horse_rest}
\end{subfigure}
\caption{The figure shows query image (left) with mapped images (right) whose feature vectors have the least Euclidean distances from the query image. The Euclidean distance increases from left to right. The KNN search performed in above images is for the best performing pretext technique, BYOL, but similar trends were observed in other techniques too. In Fig \ref{fig:KNN_Plane_main}, Fig \ref{fig:KNN_Plane_rest} we can observe that the model groups together images with blue pixels and thus tends to misclassify boats and aeroplanes. Evidence for this can be seen from Fig Fig \ref{fig:Sal_Plane} and Fig \ref{fig:Sal_Plane} where the model is classifying those images on the basis of blue regions. In Fig \ref{fig:KNN_Bird_main} and Fig \ref{fig:KNN_Bird_rest} we can observe that the mapped images appear somewhat random. This may be due to the fact that in Fig \ref{fig:Sal_Bird} the model isn't focusing on the bird but vaguely on the background. In Fig \ref{fig:KNN_Car_main} and Fig \ref{fig:KNN_Car_rest} we can see that that the model group together images of cars as well as trucks probably due to the red pixels. From Fig \ref{fig:Sal_truck} we can observe that the model focuses in the red colored pixels only. In Fig \ref{fig:KNN_Horse_main} and \ref{fig:KNN_Horse_rest} we can observe that the model has group together all horse images accurately, this may be a result of the model focusing on the horse as we can see in Fig \ref{fig:Sal_Horse}. } 
\label{fig:KNN_images}
\end{figure*}

We performed K nearest neighbors search for each architecture and technique. One common trend we observed was that, most of the techniques misclassified Aeroplanes as Ships (Fig \ref{fig:KNN_Plane_main}, Fig \ref{fig:KNN_Plane_rest}). This could be attributed to the prevalence of the blue colored pixels in both of images. Moreover this assumption is strengthened by visualization of such images through Saliency maps(Fig \ref{fig:Sal_Plane}), which showed that most models classified those images primarily based off of the blue pixels rather than the object.

Another observation is that images with substantial amount of red pixels are directly classified as Trucks by most models (Fig \ref{fig:KNN_Car_main}, Fig \ref{fig:KNN_Car_rest}). This may be due to skewed data in case of the Truck class and a general lack of other red colored objects. Subsequently, in most techniques the truck label was wrongly predicted. The Saliency map of a truck in Fig \ref{fig:Sal_truck} further supports this idea. We observe that the model classifies the image as a truck primarily based off of the red colored section.

Models were most consistently accurate while predicting horse images and least consistently accurate while predicting bird images(Fig \ref{fig:KNN_Horse_main}, Fig \ref{fig:KNN_Horse_rest}, Fig \ref{fig:KNN_Bird_main}, Fig \ref{fig:KNN_Bird_rest}). For instance, we observed from Fig \ref{fig:Sal_Horse} that the model focused on the horse with some regard for its surrounding. On the contrary, in Fig \ref{fig:Sal_Bird} we observe that there is very little focus on the bird and its classifying the image more on the basis of its surrounding. These observations were a good representation of the many other observations we made during our experiments with Saliency Maps and KNNs.

\section{Assumptions and Limitations}
\subsection{Architectures without skip connections}
After conducting most of our experiments we realized that all the architectures used for evaluation had intrinsically incorporated skip connections. Conducting experiments on architectures without skip connections would have resulted in a more comprehensive study.

\subsection{Heavier Architectures}
Our experiments have 4 light weight architectures and 1 moderately heavy architecture, but evaluation on even bigger architectures (Ex. ResNet-101) could have provided a more concrete comparison benchmark. We were unable to conduct these experiments due to the high compute requirement as well as the huge amount of time required to conduct a single experiment across all the tasks.

\subsection{Dataset}
We used a single data set for all of our experiments. While this removed the chance of inherent variation in results due to varying data it also limited our observation scope. Conducting all the experiments with a different data set could have helped in more concretely establishing all the model performances.

\section{CO2 Emission Related to Experiments}
Experiments were conducted using Google Cloud Platform in region asia-southeast1, which has a carbon efficiency of 0.42 kgCO$_2$eq/kWh. A cumulative of 250 hours of computation was performed on hardware of type Tesla P100 (TDP of 250W).

Total emissions are estimated to be 35 kgCO$_2$eq of which 100 percents were directly offset by the cloud provider.

Estimations were conducted using the Machine Learning Impact calculator presented in \cite{lacoste2019quantifying}.
% KNN

\section{Conclusion}
In conclusion we found that while having a heavier architecture does help, having a good pretext technique has a bigger impact on the performance of models. In our experiments BYOL was the best performing technique and Rotnet was consistently among the worst performing techniques. Contrastive techniques performed better than non-contrastive techniques most of the times even in the realm of lightweight architectures.

Aeroplanes ans Ships are mis-labelled as each other because the models focus on the blue pixels in the image rather than the object. Similarly Cars and Trucks are mis-labelled as each other because the model focuses on the red pixels. This could be tackled by using a more diverse data set. Bird images are the most misclassified and horse images are the most successfully classified this was attributed to weight of focus on the object and environment.

\section{Future Work}

Some things that could be explored in the future are using more diverse and balanced datasets. Experimenting on other lightweight architectures such as condensenet \cite{condensenet}, ThiNet \cite{thinet} and the others can also be explored. Computationally heavier architectures than ResNet-18, such as ResNet-50, ResNet-101, DenseNet-121 and the others can also be used to gauge the performance of various pre-text tasks. Also, incorporating more self-supervised techniques would be one of the important goals. Finally, a more diverse set of downstream tasks like object detection using the pretext task trained models as backbones would enable a fairer study of the effectiveness of architectural and method choices.

\bibliography{main_paper}
\bibliographystyle{apacite}

\end{document}